# Shortest Path Set Induced Vertex Ordering and its Application to Distributed Distance Optimal Multi-agent Formation Path Planning

Jingjin Yu

*Abstract*— For the task of moving a group of indistinguishable agents on a connected graph with unit edge lengths into an arbitrary goal formation, it was shown that distance optimal paths can be scheduled to complete with a tight convergence time guarantee [22], using a fully centralized algorithm. In this study, we show that the problem formulation in fact induces a more fundamental ordering of the vertices on the underlying graph network, which directly leads to a more intuitive scheduling algorithm that assures the same convergence time and runs faster. More importantly, this structure enables a distributed scheduling algorithm once individual paths are assigned to the agents, which was not possible before. The vertex ordering also readily extends to more general graphs - those with non-unit capacities and edge lengths - for which we again guarantee the convergence time until the desired formation is achieved.

## I. INTRODUCTION

For the task of moving a group of $n$ indistinguishable *agents* (or equivalently, *robots* or *vehicles*) on a connected graph unit length edges into an arbitrary goal formation, an efficient centralized algorithm in [22] schedules all agents from an initial formation (configuration) to a goal formation, along paths of which the total distance is minimal. Moreover, it was established that the schedule can be completed within $n + \ell - 1$ time steps ($\ell$ is the largest of the distances between all pairings of start and goal vertices), which is shown to be a tight bound.

In this paper, we show that a directed acyclic graph (DAG) induced by the initial and goal formations admits an integral ordering of the vertices on the involved paths. The ordering, which may be used to compute the distance between any two vertices on a directed path of the DAG, is unique up to an additive constant and leads to the same convergence time guarantee. This more fundamental structure provides a smooth transition from the problem formulation to the solution, which is missing from the constructive proof offered in [22]. Based on this structure, once the initial agent-target assignment is completed, the agents, using only local (up to distance 2) communication among the neighbors, can achieve the desired formation within identical convergence time. To the best of our knowledge, this work provides the first multi-agent formation path planning algorithm that is both truly distance optimal and partially distributed, along with a tight convergence time guarantee (keep in mind that global distance optimality is not possible without direct or indirect global communication, implying that a fully distributed planning algorithm is out of the question here). We implemented the algorithm which is accessible online (see Section V). As we will see, the ordering also allows easy extension of the results to graphs with edges having arbitrary integer lengths and non-unit capacities (i.e., more than one agent may be traveling on the same edge at a given instant). This paragraph captures the main contributions of our study.

When it comes to problems on *formation*, two subproblems come up. One of them is on the topic of formation control, which focuses on maintaining a formation of a group of vehicles; a desired formation, in these research, may be important for inter-vehicle communication or for maximizing certain utility functions [4], [16], [24]. Graph theoretic approaches are quite popular here, probably because vehicles and inter-vehicle constraints can be represented naturally with vertices and edges of graphs. The second subproblem put more emphasis on how to achieve a desired formation (as opposed to stabilizing around a given formation) [3], [5], [7], [8], [11], [21], [12], [15], [19], [20], which is the problem we address in this paper. On research that appears most related to our problem, a discrete grid abstraction model for formation control was studied in [10]. To plan the paths, a three-step process was used in [10]: 1) Target assignment, 2) Path allocation, 3) Trajectory scheduling. Although it was shown that the process always terminates, no characterization of solution complexity was offered. In contrast, we provide efficient algorithms for solving a strictly more general class of problems with optimality assurance.

Generalizing the notion of formation to include multiple agents trying to agree on some common goal leads to the problem of consensus/rendezvous. This more general problem has remained a central research topic in control theory and robotics; see, e.g., [1], [2], [6], [13], [14], [16], [17], [18], [23], to list a few. An early account of the rendezvous problem, as a form of formation control, appeared in [1], in which algorithmic solutions are provided for agents with limited range sensing capabilities. An $n$-dimensional rendezvous problem was approached via proximity graphs in [2]. For the consensus problem it is shown that averaging the behavior of close neighbors causes all agents to converge to the same behavior eventually [6]. We point out that, although this paper works with disjoint initial and goal vertex sets of $n$ distinct elements each, the presented results can be easily generalized to any number of goal vertices between 1 and $n$, thus covering additional problems such as multi-agent rendezvous.

The rest of the paper is organized as follows. Section

Jingjin Yu is with the Department of Electrical and Computer Engineering, University of Illinois at Urbana-Champaign, Urbana, IL 61801 USA. E-mail: jyu18@uiuc.edu. The author was supported in part by DARPA SToMP grant HR0011-05-1-0008, NSF grants 0904501 (IIS Robotics) and 1035345 (Cyberphysical Systems), and MURI/ONR grant N00014-09-1-1052.

II provides the problem formulation, an example, and its solution. Section III constructively proves the existence of the afore mentioned vertex ordering on the induced DAG. Section IV shows an application of the vertex ordering in scheduling a set of distance optimal paths for the agents. Section V then shows the scheduling algorithm can be easily turned into a distributed one. We generalize the graph to have integer edge lengths and capacities in Section VI and conclude in Section VII.

## II. FORMATION PATH PLANNING ON GRAPHS

Let $G = (V,E)$ be a connected, undirected, simple graph, in which $V = \{v_i\}$ is its vertex set and $E = \{(v_i, v_j)\}$ is its edge set. Let $A = \{a_1, \ldots, a_n\}$ be $n$ agents that move with unit speeds along the edges of $G$, with initial and goal vertices on $G$ specified by the injective maps $x_I, x_G : A \to V$, respectively. The set $A$ is effectively an index set. For convenience, $V, E$ also denote cardinalities of the sets $V, E$, respectively. Let $\sigma$ be a permutation that acts on the elements of $x_G$, $(\sigma \circ x_G)$ is a map that defines a possible goal vertex assignment (a target formation).

A *scheduled path* is a map $p_i : \mathbb{Z}^+ \to V$, in which $\mathbb{Z}^+ := \mathbb{N} \cup \{0\}$. Intuitively, the domain of the paths is discrete time steps. A scheduled path $p_i$ is *feasible* for a single agent $a_i$ if it satisfies the following properties: 1) $p_i(0) = x_I(a_i)$. 2) For each $i$, there exists a smallest $k_{\min} \in \mathbb{Z}^+$ such that $p_i(k_{\min}) = (\sigma \circ x_G)(a_i)$ for some fixed $\sigma$ (i.e., same $\sigma$ for all $1 \le i \le n$). That is, the end point of the path $p_i$ is some unique goal vertex. 3) For any $k \ge k_{\min}$, $p_i(k) \equiv (\sigma \circ x_G)(a_i)$. 4) For any $0 \le k < k_{\min}$, $(p_i(k), p_i(k+1)) \in E$ or $p_i(k) = p_i(k+1)$. We say that two paths $p_i, p_j$ are in *collision* if there exists $k \in \mathbb{Z}^+$ such that $p_i(k) = p_j(k)$ (*meet*, or collision on a vertex) or $(p_i(k), p_i(k+1)) = (p_j(k+1), p_j(k))$ (*head-on*, or collision on an edge). If $p(k) = p(k+1)$, the agent stays at vertex $p(k)$ between the time steps $k$ and $k+1$.

**Problem 1 (Formation Control on Graphs)** *Given a 4-tuple $(G, A, x_I, x_G)$, find a set of paths $P = \{p_1, \ldots, p_n\}$ and a fixed $\sigma$ such that $p_i$'s are feasible paths for respective agents $a_i$'s for this $\sigma$ and no two paths $p_i, p_j$ are in collision.*

Note that in the definition above, we have implicitly assumed that edges of $G$ have unit lengths and capacities. That is, an edge takes unit time for an agent to cross and no two agents can be on an edge at the same time. This implicit assumption is used throughout Section III-V and relaxed in Section VI.

To familiarize readers with the problem and its solution, look at the example in Fig. 1. The underlying graph $G$ is a $6 \times 7$ grid with holes. Assigning the top left corner coordinates $(0,0)$ and bottom right coordinates $(6,5)$, $x_I(A) = \{(0, i-1)\}, x_G(A) = \{(6, i-1)\}, 1 \le i \le 6$. That is, we want to move the agents from left to right. A solution to this problem that is *distance optimal* is given in Table I, corresponding to a *schedule* of the multi-colored paths in Fig. 1. Here, *distance optimality* seeks to minimize the total path lengths of all agents. Each main entry of the table designates the coordinates of the vertex an agent should be staying at the given time step.

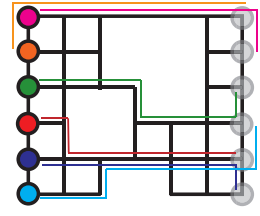

Fig. 1. A $6 \times 7$ grid with some vertices removed. The colored discs on the left represent the initial formation and the gray discs represent the goal formation. The colored paths represent the paths (not yet scheduled to avoid collision).

TABLE I

| Agent | Time Step | | | | | | | | |
|---|---|---|---|---|---|---|---|---|---|
| | 0 | 1 | 2 | 3 | 4 | 5 | 6 | 7 | 8 |
| 1 | 0,0 | 1,0 | 2,0 | 3,0 | 4,0 | 5,0 | 6,0 | 6,1 | 6,1 |
| 2 | 0,1 | 0,0 | 1,0 | 2,0 | 3,0 | 4,0 | 5,0 | 6,0 | 6,0 |
| 3 | 0,2 | 1,2 | 2,2 | 3,2 | 3,3 | 4,3 | 5,3 | 6,3 | 6,2 |
| 4 | 0,3 | 1,3 | 1,4 | 1,4 | 2,4 | 3,4 | 4,4 | 5,4 | 6,4 |
| 5 | 0,4 | 1,4 | 2,4 | 3,4 | 4,4 | 5,4 | 6,4 | 6,5 | 6,5 |
| 6 | 0,5 | 1,5 | 2,5 | 2,4 | 3,4 | 4,4 | 5,4 | 6,4 | 6,3 |

## III. FORMATION INDUCED VERTEX ORDERING

**Algorithm 1** PLANSHORTESTPATHSET

**Input:** $G, A, x_I, x_G$ as described in Problem 1
**Output:** $Q = \{q_1, \ldots, q_n\}$

1: **for** each $u_i \in x_I(A)$ **do**
2:   run Dijkstra's algorithm to get shortest paths $q_{ij}$ for all $(u_i, v_j)$'s such that $v_j \in x_G(A)$
3: **end for**
4: run minimum weighted bipartite matching algorithm on the above set of $n^2$ paths to get a path set $Q$.
5: **return** $Q$

Given $x_I$ and $x_G$, it is relatively straightforward to obtain a *unscheduled* path set $Q = \{q_1, \ldots, q_n\}$ in which $q_i$ is a sequence of vertices such that each is adjacent to the next in the sequence (we use $Q$ to distinguish these paths from the scheduled paths, which we denote as $P$). Such a procedure, generally known as the Hungarian algorithm [9], is outlined in Algorithm 1 (from [22]). Let $head(q_i), tail(q_i)$, and $len(q_i)$ denote the start vertex, end vertex, and length of $q_i$, respectively. The path set $Q$ returned from Algorithm 1 has the following properties:

**Property 2** *For all $1 \le i \le n$, $head(q_i) \in x_I(A)$ and $tail(q_i) \in x_G(A)$. For any two paths $q_i, q_j$, $head(q_i) \ne head(q_j)$ and $tail(q_i) \ne tail(q_j)$.*

**Property 3** *Each path $q_i$ is a shortest path between $head(q_i)$ and $tail(q_i)$ on $G$.*

**Property 4** *The total length of the path set Q is minimal.*

Constructively guaranteed by Algorithm 1, Properties 2 and 3 ensure that the initial and goal vertices are paired up using shortest paths. Property 4 requires the total length of these paths to be minimal. From now on, $Q$ is always assumed to be a path set satisfying properties 2-4. It is not hard to see that Property 4 implies the following [22]:

**Property 5** *If we orient the edges of every path $q_i \in Q$ from $head(q_i)$ to $tail(q_i)$, no two paths share a common edge oriented in different directions.*

With a slight abuse of notation, let $V(\cdot), E(\cdot)$ denote the vertex set and undirected edge set of the input argument, which can be either a path, $q_i$, or a set of paths, such as $Q$. We define an *intersection* between two paths as a maximal consecutive sequence of vertices and edges common to the two paths. Property 5 is a special case of a more general structure of the path set $Q$ [22].

**Proposition 6** *The path set $Q$ induces a directed acyclic graph (DAG) structure on $E(Q)$.*

With Proposition 6, it is then possible to bound the total number of time steps necessary to schedule the path set $Q$. Somewhat surprisingly, the DAG structure on $Q$ has an even stronger order property that does not hold in general for DAGs; this is where the contribution of this paper starts. To state this property, some definitions are needed for describing relationships between paths. Recall that two paths *intersect* (a symmetric relationship) if they share some common vertices or edges. Two paths $q_i, q_j$ are *linked* (again a symmetric relationship) if either $q_i, q_j$ intersect or both $q_i, q_j$ are *linked* to some $q_k$. A *cluster* $Q_c$ is a set of paths such that every pair of paths $q_i, q_j \in Q_c$ are linked. A path cluster $Q_c$ is a *maximal* cluster of $Q$ if $Q_c$ is a cluster and no other path $q_i \in Q \backslash Q_c$ is linked to a path $q_j \in Q_c$. For each path $q_i \in Q$, define a *distance value function*, $d_i : V \to \mathbb{Z}^+$, such that for $u \in V(q_i)$,

$$d_i(u) = \begin{cases} 0 & u = head(q_i), \\ dist(head(q_i), u) & \text{otherwise}, \end{cases} \quad (1)$$

in which $dist(u,v)$ denotes the shortest distance between $u,v$ on the graph $G$. Distance value functions can be defined similarly for an arbitrary set of vertices. Given the generalized definition, we say that one distance value function, $d'$, *respects* another one, $d$, if $d'$ is defined for all of $d$'s domain and for any $u,v$ on which $d$ is defined,

$$d'(u) - d'(v) = d(u) - d(v). \quad (2)$$

In an unscheduled path set $Q$, for any two paths $q_i, q_j$ that intersect, it is not hard to construct a distance value function that respects both $q_i$ and $q_j$.

**Lemma 7** *If a vertex $u^*$ belongs to the intersection of two paths $q_i, q_j \in Q$, then the value function*

$$d_c(u) = \begin{cases} d_i(u) & u \in V(q_i), \\ d_c(u^*) + d_j(u) - d_j(u^*) & u \in V(q_j), \end{cases} \quad (3)$$

*respects both $d_i$ and $d_j$.*

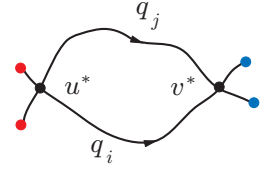

Fig. 2. Two intersections between two paths.

PROOF. By definition, $d_c$ respects $d_i$. If there is a single intersection (or common segment) between $q_i$ and $q_j$, then $d_c$ respects $d_j$ since by Property 5, $q_i$ and $q_j$ cannot have edges oriented differently. If not, let $v^*$ be another intersection point of $q_i, q_j$ such that that segments of $q_i, q_j$ between $u^*, v^*$ differ (see Fig. 2). Denote the segment between $u^*, v^*$ as $\omega_i, \omega_j$ for $q_i, q_j$, respectively. Note that $\omega_i, \omega_j$ must be oriented the same way between $u^*, v^*$ by Proposition 6. We want to show that $d_i(v^*) = d_c(v^*)$, or

$$\begin{aligned} d_i(v^*) &= d_c(v^*) = d_c(u^*) + d_j(v^*) - d_j(u^*) \\ &= d_i(u^*) + d_j(v^*) - d_j(u^*) \\ \Leftrightarrow \quad d_i(v^*) - d_i(u^*) &= d_j(v^*) - d_j(u^*) \\ \Leftrightarrow \quad len(\omega_i) &= len(\omega_j) \end{aligned} \quad (4)$$

The last equation of (4) holds since otherwise, for example $len(\omega_i) < len(\omega_j)$, then both path should have taken $\omega_i$. Since $v^*$ is arbitrary, we have shown that $d_c$ respects both $q_i, q_j$. □

We now show that (7) extends $d_c$ to a path cluster.

**Theorem 8** *Given a path cluster, $Q_c = \{q_1, \ldots, q_m\} \subset Q$, there exists a distance value function $d_c : V(Q_c) \to \mathbb{Z}^+$, such that $d_c$ respects $d_i$ for all $1 \leq i \leq m$.*

PROOF. Lemma 7 proves the case for any path cluster with no more than two paths. To complete the proof, assuming that given a $d_c$ that respected a sub cluster $\{q_1, \ldots, q_{k-1}\} \subset Q_c$, we show that the recursively definition

$$d_c(u) = \begin{cases} d_c(u) & u \in V(\{q_1, \ldots, q_{k-1}\}), \\ d_c(u^*) + d_k(u) - d_k(u^*) & u \in V(q_k), \end{cases} \quad (5)$$

extends $d_c$ so that it respects $d_k$ for a path $q_k$ that intersects paths in $\{q_1, \ldots, q_{k-1}\}$ at some vertex $u^*$. The case is trivial if $q_k$ intersects paths in $\{q_1, \ldots, q_{k-1}\}$ only once or $q_k$ intersects a single path $q_i, 1 \leq i \leq k-1$ more than once.

The leftover case is that $q_k$ intersects $q_i, q_j, 1 \leq i < j \leq k-1$, at $u^*, v^*$, respectively (the order of $i, j$ does not matter), with $u^*, v^*$ being $q_k$'s earliest two intersections with paths in $\{q_1, \ldots, q_{k-1}\}$. We may assume that the intersection has the general structure illustrated in Fig. 3 (in the figure there are 7 paths on the undirected cycle; there could be fewer or more paths). We assume a single intersection (see Fig. 3) between each pair of paths since we have shown that multiple intersections between the same pair of paths do not affect $d_c$ values. The other possibility for the assumption to be violated is that some path, for example $q_i$, takes a different orientation from that shown in Fig. 3. When this happens, it is always possible to update $\{q_1, \ldots, q_k\}$ to a new path set such that fewer paths are on the undirected cycle. We show how to handle this exception using an example in which $q_i$ takes a different orientation (see Fig. 4); all other cases are similar.

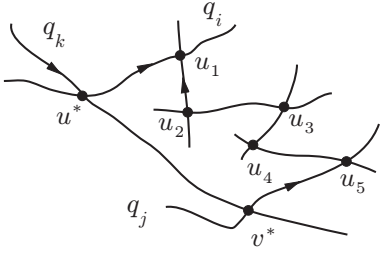

Fig. 3. A typical multiple path intersection.

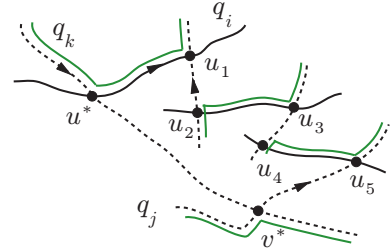

Fig. 5. Augmented paths. The dotted paths are the old paths, replaced by the green ones. Note that the two sets of paths (green and dotted ones) have the same initial and goal vertex sets.

For the case depicted in Fig. 4, we update $q_i, q_k$ to $q'_i, q'_k$ (green paths in Fig. 4). It is straightforward to verify that Properties 2-4 are not violated. Furthermore, $d_c$ as obtained for $\{q_1, \ldots, q_{k-1}\}$ readily extends to $q'_i$ via $d_c(u^*)$ since the segment of $q_k$ before $u^*$ does not intersect $\{q_1, \ldots, q_{k-1}\}$ by assumption. We are left with the same general setting we begin with, except that there is one fewer path ($q'_i$) on the cycle.

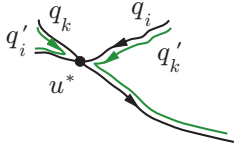

Fig. 4. A scenario in which $q_i$ has a different orientation compared to Fig. 3.

Returning to the case illustrated in Fig. 3, on the undirected cycle formed by intersecting paths, there is always an even number of paths between $u^*, v^*$ besides $q_k$ (possibly after applying above path switching procedure many times). Let this even number be $2b$ and the $2b$ paths intersect at $u_1, \ldots, u_{2b-1}$ (Fig. 3 shows the case where $b = 3$). Also let $u^* \equiv u_0, v^* \equiv u_{2b}$. To show that (5) extends $d_c$ to $d_k$, we need to show $d_k(v^*) - d_k(u^*) = d_c(v^*) - d_c(u^*)$. If this is not true, without loss of generality, assume that $d_k(v^*) - d_k(u^*) > d_c(v^*) - d_c(u^*)$. For this case, we update segments of the paths as shown in Fig. 5. The update gives us a net gain of path length

$$\begin{aligned}
&-dist(u^*, v^*) + \sum_{i=0}^{b-1} dist(u_{2i}, u_{2i+1}) - \sum_{i=1}^{b} dist(u_{2i-1}, u_{2i}) \\
&= d_k(u^*) - d_k(v^*) + \sum_{i=0}^{b-1}(d_c(u_{2i+1}) - d_c(u_{2i})) \\
&\quad + \sum_{i=1}^{b}(d_c(u_{2i-1}) - d_c(u_{2i})) \\
&= d_k(u^*) - d_k(v^*) + d_c(v^*) - d_c(u^*) < 0,
\end{aligned}$$

which contradicts Property 4. We conclude that (5) indeed extends $d_c$ to respect $d_k$. $\square$

## IV. A SIMPLE PATH SCHEDULING ALGORITHM BASED ON THE INDUCED VERTEX ORDERING

Assuming that a *time optimal* schedule seeks to minimize the time it takes the last agent to reach its goal, the following was established in [22]:

**Lemma 9** *Distance optimality and time optimality for Problem 2 cannot be simultaneously satisfied.*

Furthermore, let $\ell$ be the largest pairwise distance between a member of $x_I(A)$ and a member of $x_G(A)$,

$$\ell = \max_{\forall u \in x_I(A), v \in x_G(A)} dist(u, v). \quad (6)$$

It was also shown in [22] that $n + \ell - 1$ time steps is sometimes necessary to schedule a shortest path set $Q$. It was then shown that an unscheduled path set $Q$ can be turned into a scheduled path set $P$ with a maximum of $n + \ell - 1$ time steps, providing a distance optimal schedule with a tight scheduling time bound. We now show that the vertex ordering induced by $x_G$ leads to a scheduling algorithm with the same guarantees on the scheduled paths' qualities. The new algorithm has a better running time of $O(nV \log n)$ and is easy to understand; it is not clear though, from a first look, that it should provide the said convergence time guarantee.

By Theorem 8, each maximal path cluster $Q_c \subset Q$ can be assigned a distance value function $d_c$ that respects every $q_i \in Q_c$. Since these individual $d_c$'s have no common domain, they can be combined to give a global $d_c$ (for a fixed $Q$).

Assuming such a $d_c$, which can be obtained easily using (5). Before scheduling the path set $Q$, we introduce a subroutine to handle the scenario illustrated in Fig. 6. In the figure, $Q = \{q_1, q_2\}$ with $head(q_i) = u_i, tail(q_i) = v_i$ for $i = 1, 2$. This path set cannot be scheduled as is, since $q_1$ is in the way

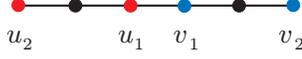

$u_2 \quad\quad u_1 \ v_1 \quad\quad v_2$

Fig. 6. A path set $Q$ that cannot be scheduled without modification.

of $q_2$. However, as agent $a_1$ reaches $v_1$, we can dynamically switch the goals of $q_1, q_2$. Note that the path set after this update still satisfies Properties 2-4. For paths $q_i, q_j$, denote this path switching subroutine $switch(q_i, q_j)$.

**Algorithm 2** SCHEDULESHORTESTPATHS

**Input:** $G, Q, d_c$
**Output:** scheduled paths, $P = \{p_1, \ldots, p_n\}$

1: let $p_i(0) = head(q_i)$ for all $1 \leq i \leq n$
2: let $v_i = next(q_i, head(q_i))$ for all applicable $q_i \in Q$
3: let $t = 1$
4: **while** some $v_i$ is not fully scheduled **do**
5:   **while** some $p_i(t)$ is not set for the current $t$ **do**
6:     pick a candidate path $q_i$ with largest $d_c(v_i)$
7:     **if** $v_i$ is not the same as any $p_j(t)$ already assigned **then**
8:       $p_i(t) = v_i$
9:       $v_i = next(q_i, v_i)$ if $q_i$ is not fully scheduled
10:       **if** $v_i == tail(q_i)$ and $v_i$ falls on some $q_j$ such that $q_j$ has yet to reach $v_i$ **then**
11:         $switch(q_i, q_j)$
12:       **end if**
13:     **else**
14:       $p_i(t) = p_i(t-1)$
15:     **end if**
16:   **end while**
17:   $t = t + 1$
18: **end while**
19: **return** $P = \{p_1, \ldots, p_n\}$

The path scheduling subroutine is outlined in Algorithm 2, in which the routine $next(q_i, v)$ returns the next vertex of path $q_i$ after vertex $v$. A path $q_i$ is *fully scheduled* if $tail(q_i)$ is assigned to $p_i(t)$ for some $t$. The scheduling routine never considers two paths $q_i, q_j$ running in opposite directions since Property 5 excludes such cases. Essentially, the scheduling algorithm let all paths from $Q$ take their respective courses simultaneously. Whenever two paths are competing for going to the same vertex, an arbitrary path is picked to go and the other one to stay put. With the $switch(\cdot, \cdot)$ subroutine to guarantee that no deadlock can occur, it is straightforward to see that the process must converge since at each $t$, at least one agent will make progress toward its goal. That is,

**Proposition 10** *Algorithm 2 terminates in finite time.*

Denote the total path length of $Q$ as $\ell_Q$, then the convergence time (the time it takes for the formation to be completed) is no more than $\ell_Q$. However, as we have mentioned, Algorithm 2 provides a much stronger guarantee, as Theorem 11 will show. we apologize for the somewhat long proof but it seems more appropriate to have a long proof in this case than to split it into lemmas.

**Theorem 11** *Algorithm 2 provides a schedule that takes at most $n + \ell - 1$ time steps to complete.*

PROOF. We constructively prove the theorem starting with a path set $Q$. Let $Q_c = \{q_1, \ldots, q_m\}$ be an arbitrary maximal cluster of $Q$, it is clear that the schedule of $Q_c$ is not affected by any other path $Q \backslash Q_c$; hence, we only need to prove the claim for $Q_c$. Moreover, we only need to prove the bound for the special case in which the routine $switch(\cdot, \cdot)$ is never invoked, since we can effectively consider the "dynamic switching" all happen at time step $t = 0$. Note that it is easy to see, once we go through the proof, that it also holds if dynamic switching were performed on the fly.

We want to schedule all agents, $a_1, \ldots, a_m$, along $q_1, \ldots, q_m$, respectively, starting at $t = 0$. Before starting, we create a list of numbers, $H$, indexed by possible $d_c$ values (as constructed in the proof of Theorem 8) in decreasing order. Since the cluster $Q_c$ is finite, $H$ is also finite. An entry of this list, $h_d$, is the number of agents whose current locations have a $d_c$ value of $d$. A list $H$ may look like

| $d$ | : | 5 | 4 | 3 | 2 | 1 | 0 | $-1$ | $-2$ | $-3$ | $-4$ | ... |
|---|---|---|---|---|---|---|---|---|---|---|---|---|
| $h_d$ | : | 0 | 0 | 3 | 0 | 2 | 1 | 0 | 0 | 0 | 2 | ... |

We note that for a fixed time step $t$, the only importance of the index $d$ in list $H$ is that it specifies the relative order of agents. In the above $H$, for example, $h_3 = 3$, as the first non-zero entry, means that there are 3 agents as the "front runners", followed by next non-zero entry $h_1 = 1$, suggesting that there is 1 agent two steps behind. From this observation, we may negate the index $d$ and at each time step $t$, align $h_1$ with the first non-zero entry of $H$. The above $H$ then becomes

| $d$ | : | $-1$ | 0 | 1 | 2 | 3 | 4 | 5 | 6 | 7 | 8 |
|---|---|---|---|---|---|---|---|---|---|---|---|
| $h_d$ | : | 0 | 0 | 3 | 0 | 1 | 2 | 0 | 0 | 0 | 2 |

We can also remove the leading (and trailing) zero entries

| $d$ | : | 1 | 2 | 3 | 4 | 5 | 6 | 7 | 8 |
|---|---|---|---|---|---|---|---|---|---|
| $h_d$ | : | 3 | 0 | 1 | 2 | 0 | 0 | 0 | 2 |

At this point, we partition the list into one or more sublists as follows. Starting at $d = 0$, we look at the sum of first $k$ terms of $H$,

$$S_k = \sum_{d=1}^{k} h_d. \quad (7)$$

If it ever happens for some $k$, starting at 1, that $S_k = k$, we group these $k$ terms of $H$ into a sublist and work with it. We call these sublists *contiguous* sublists. Applying the partition procedure to $H$ above, the first contiguous sublist, $H'$, is

| $d$ | : | 1 | 2 | 3 | 4 | 5 | 6 |
|---|---|---|---|---|---|---|---|
| $h'_d$ | : | 3 | 0 | 1 | 2 | 0 | 0 |

Now let us consider how the sublist may change after we let all agents start moving towards their respective goals. For the first group of 3 agents, at least one of them can move one step closer to its goal and at most two of them may not make any progress. The worst case happens in a situation illustrated in Fig. 7. When conflict like this happens, we pick a random agent to advance. Suppose the worst case happens, we update to have $h'_0 = 1, h'_1 = 2$.

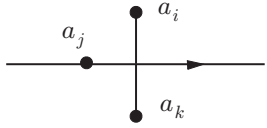

Fig. 7. A three-way conflict between three agents, in which case we pick any agent to go ahead and let the other two wait.

For the rest of the non-zero entries of $H'$, if it is preceded by a zero entry, it means that there are no agents with $d_c$ values exactly one larger than this group of agents. Hence, at least one agent from this group of agents can make one step progress towards its goal. When this is applied to $h'_3$, which has a value 1, we update the sublist entries as $h'_2 = 1, h'_3 = 0$. After these two steps, the sublist $H'$, still being processed between time steps, is

$$d \;\; : \;\; 0 \;\; 1 \;\; 2 \;\; 3 \;\; 4 \;\; 5 \;\; 6$$
$$h'_d \;\; : \;\; 1 \;\; 2 \;\; 1 \;\; 0 \;\; 2 \;\; 0 \;\; 0$$

Observe that the next non-zero entry, $h'_4 = 2$, now has a preceding zero entry. Assume that only one agent advances, we update the entries to $h'_3 = 1, h'_4 = 1$. At the end of $t = 1$, the updates give us $H'$ as

$$d \;\; : \;\; 0 \;\; 1 \;\; 2 \;\; 3 \;\; 4 \;\; 5 \;\; 6$$
$$h'_d \;\; : \;\; 1 \;\; 2 \;\; 1 \;\; 1 \;\; 1 \;\; 0 \;\; 0$$

We can adjust $d$ to get

$$d \;\; : \;\; 1 \;\; 2 \;\; 3 \;\; 4 \;\; 5 \;\; 6$$
$$h'_d \;\; : \;\; 1 \;\; 2 \;\; 1 \;\; 1 \;\; 1 \;\; 0$$

What does the entries of $H'$ mean after each update? The entry $h'_1$ represents the number of agents (paths) that never waited for others. Similarly, the entry $h'_i$ represents the number of agents (paths) that never waited for more than $i - 1$ steps. Note that the leading entry is $h'_1 = 1$. Whenever the leading entry becomes 1, the associated agent/path can no longer have any conflict with any other agent/path. That is, it has no more interaction with the rest of the agents. We claim that throughout the updates, at least $i$ agents never waited more than $i - 1$ steps, which is easily verifiable via induction (we omit the details due to its length and irrelevance to the rest of the paper). The worst case happens when $H'$ becomes all 1's as

$$d \;\; : \;\; 1 \;\; 2 \;\; 3 \;\; 4 \;\; 5 \;\; 6$$
$$h'_d \;\; : \;\; 1 \;\; 1 \;\; 1 \;\; 1 \;\; 1 \;\; 1$$

In any case, by the above claim, agents from a contiguous sublist cannot "spill over" to the next contiguous list. If we apply what we have done to every contiguous sublist, the claim that at least $i$ agents never waited more than $i - 1$ steps holds for all agents moving on the cluster $Q_c$. Since no agent travels a length more than $\ell$, We have proved the claim of the theorem. □

## V. A Distributed Scheduling Algorithm

Looking closely at the constructive proof of Theorem 11, it is not hard to observe the following: 1. Different path clusters can be scheduled independently; 2. Within each path cluster, an agent only needs to be aware of its neighbors within a distance of 2 to take appropriate actions. These observations imply the possibility of a partially distributed planning algorithm that yields shortest total path length: Once agent-target assignment is done, it seems that global coordination is not required to *schedule* these agents. Since local communication is often more reliable and easy to implement, such a scheduling algorithm is more desirable in general. In this section, we provide a local communication protocol which leads to a distributed scheduling algorithm, again with a convergence time of $n + \ell - 1$. A common clock is assumed. Since the algorithm is essentially a distributed version of Algorithm 2, we omit the pseudocode.

Assuming each agent is assigned a path, we will schedule them along these paths and possibly update their goals (targets) on the fly. Recall that with Property 5, we only need to worry about two agents occupying the same vertex at a given time step. This splits into two cases: 1. Two agents want to move to the same vertex in one time step, and 2. One agent moves to a vertex while another agent is staying there. We now give a communication protocol, including a forward communication phase and a backward communication phase at each time step, that handles both cases.

**Schedule 12 (Distributed Transfer Schedule)** *Repeat the following two communication phases until the desired formation is complete.* **Forward communication phase**. *Assume that an agent $a_i$ is located on $v_i$ and wants to move to $v_{i+1}$. Agent $a_i$ first checks whether $v_{i+1}$ is occupied by some other agent $a_j$ and if it is, notifies $a_j$ of its intention and waits for $a_j$'s response. At this point, $a_j$ will check whether it is already at its goal and if it is, switch its goal with $a_i$ ($a_j$ will also redo its forward communication phase if it already did). If no agent is occupying $a_j$, $a_i$ then looks for agents that also want to go to $v_{i+1}$. If there are, one agent is randomly picked to go to $v_{i+1}$ in the next time step. Alternatively, we could deterministically pick an agent (e.g. based on identities of the vertices occupied by the agents). Other agents wanting to go to $v_{i+1}$ then must wait one time step. Since we are dealing a finite number of agents and there are no cycles on a DAG, the forward communication will stop after at most $O(n)$ messages, each with a size of $O(\log V)$.* **Backward communication phase**. *Next, an agent that has received requests from a following agent needs to respond back. Let such two adjacent agents be $a_i$ and $a_j$, occupying $v_i$, $v_{i+1}$, respectively, with $a_i$ wanting to go to $v_{i+1}$. There are two subcases: 1. If $a_j$ will move, then it will*

*notify that $a_i$ it may go ahead and move to $v_{i+1}$. If $a_j$ gets multiple requests to occupy $v_{i+1}$ then a randomly agent is selected to proceed (again, this can be made deterministic). 2. If $a_j$ cannot move because another agent tells it so, then it simply relay that message backward. Clearly, the backward communication will stop after at most $O(n)$ messages, each with a size of $O(\log V)$.*

The distributed scheduling algorithm have a similar running time compared with the centralized version. Using Theorem 11 and in particular the fact that at least $i$ agents never need to wait more than $i-1$ time steps, the following Proposition is immediate.

**Proposition 13** *Schedule 12 transfers all agents to achieve the desired formation in $O(n+\ell-1)$ time steps.*

The scheduling algorithm is fairly simple to implement, as we did in Java. For visualizing the result, a Java applet, accessible online[1], was created to illustrate the target assignment and scheduling process. For more details about the simulation, see the webpage given in the footnote. A snapshot of a running session is provided in Fig. 8. We do not provide computational evaluation here since the overall algorithm has similar running time as the algorithm from [22]. Readers interested in computational time on large instances may refer to that paper for more details.

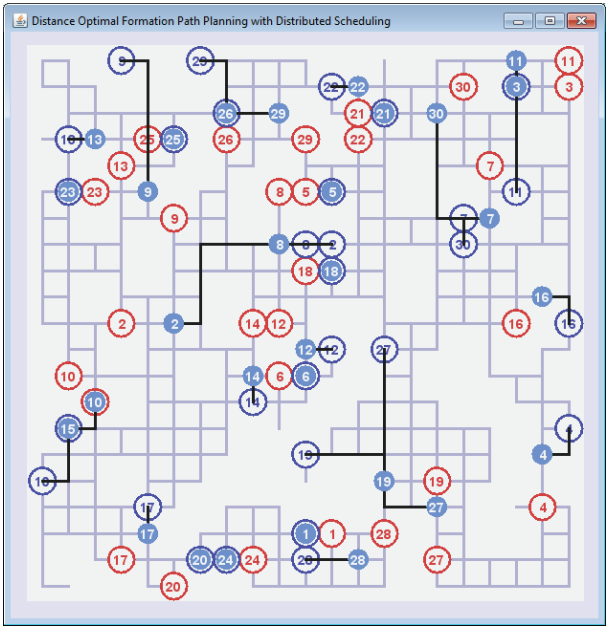

Fig. 8. A simulation capture. The red/blue circles and numbers are the start/goal locations (already assigned to have shortest total distance). The light blue solid discs represent the agents. The bold black lines are the paths yet to be completed.

---
[1]http://msl.cs.uiuc.edu/~jyu18/pe/distr-form.html (a Java plugin of version 6 or higher is required).

## VI. INTEGER EDGE LENGTHS AND CAPACITIES

So far we have assumed that we work with a graph $G$ with unit edge lengths and capacities. That is, an edge takes a unit of time to cross and can hold one agent at a time. We now relax this assumption to allow non-unit edge lengths and capacities. Formally, let $d, c: E \to \mathbb{Z}^+$ be the edge length map and edge capacity map, respectively. We assume that for any $e \in E, d(e) \geq c(e)$, which is generally true for physical robots with non-negligible sizes (up to a multiplicative constant). The main goal of this section is to extend the results from previous sections under this setup. Note that the definition of *scheduled paths* and *feasible paths* from Section II need to be updated since it may take multiple time steps for an agent to cross an edge. Thus, a scheduled path $p_i$ becomes a partial map as it may be undefined for some time steps. We omit formal descriptions of these required updates since they are intuitive but lengthy to state.

It is clear that Algorithm 1 is insensitive to edge length. Therefore, the algorithm again produces an unscheduled path set $Q$ satisfying Properties 2-5. Moreover, all results from Section III continues to hold with edge lengths that are not all ones. On the other hand, scheduling the path set $Q$ becomes slightly trickier, since depending on edge capacities, one or more agent may be on the same edge during within one time step. To simplify the analysis, we look at two extreme cases: 1. For all $e \in E, c(e) = d(e)$. 2. For all $e \in E, c(e) \equiv 1$. The first case models scenarios that allow bumper to bumper road traffic. This case is easy to handle, due to the following observation: By subdividing each edge $e \in E$ into $d(e)$ edges of unit length, we obtain a new graph $G$ with unit edge length and capacity. We turn our attention to the second case, which models bottleneck edges such as a long and thin bridge. First we establish a lower bound.

**Lemma 14** *Assume $\forall e \in E, c(e) \equiv 1$ and let $d_{\max} = \max_{e \in E} d(e)$. Then $\ell + (n-1)d_{\max}$ time steps is necessary to schedule $n$ agents along a shortest path set $Q$.*

PROOF. In the instance of Problem 1 shown in Fig. 9, assume that all edges have the same length $d$; hence, $d_{\max} = d$. The graph $G$ is two stars with their centers connected by a single path; the red vertices form $x_I(A)$ and the blue ones $x_G(A)$. It is clear that all red vertices are of distance $\ell$ to all blue vertices. Given this problem instance, all agents must go through the path $uv\ldots xy$ sequentially. To optimize arrival time, the first agent (say $a_1$) to reach goal must visit $u$ at $t = d_{\max}$. Consequently, $a_1$ cannot reach $v$ earlier than $t = 2d_{\max}$. This implies that no other agent can head to $v$ from $u$ before $t = 2d_{\max}$, due to the unit edge capacity constraint. Via simple induction, the last agent arriving at $u$ cannot leave it before $t = nd_{\max}$. Therefore, it cannot arrive earlier than $t = \ell + (n-1)d_{\max}$. □

If we pretend that all edges have the same length $d_{\max}$, Algorithm 2 can be easily extended to schedule a shortest path set $Q$. Clearly, this provides an overestimate of the total time it takes to schedule $Q$. Since no agent is delayed

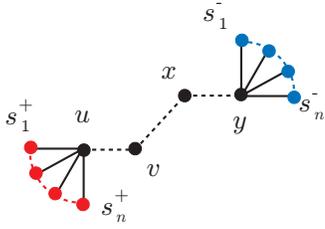

Fig. 9. An instance of Problem 1 for demonstrating the necessity claim of Lemma 14.

more than $(n-1)d_{\max}$ time steps, the following corollary to Theorem 11 is immediate.

**Corollary 15** *Assume $\forall e \in E, c(e) \equiv 1$ and let $d_{\max} = \max_{e \in E} d(e)$. Algorithm 2 schedules a shortest path set $Q$ such that the scheduled path set requires at most $\ell + (n-1)d_{\max}$ time steps to complete.*

Thus, the time bound $\ell + (n-1)d_{\max}$ is tight for the unit edge capacity case. Combining the two extreme cases together, we have the following conclusion.

**Theorem 16** *For the extension of Problem 1 with integer edge lengths and capacities in which $1 \le c(e) \le d(e)$ for all $e \in E$, the time bound $\ell + (n-1)d_{\max}$ is sufficient and necessary to schedule $n$ agents along a shortest path set $Q$.*

Straightforward complexity analysis shows that for integer edge lengths and capacities, the running time of the entire algorithm becomes $O(nV^2 + nVd_{\max})$.

## VII. Conclusion and Future Work

In this paper, for the multi-agent formation path planning problem on graphs, we showed the existence of a vertex ordering structure induced by the initial and goal formations, which in turn admits a simple and natural scheduling algorithm for coordinating the shortest paths amongst the indistinguishable agents with a tight convergence time guarantee. Furthermore, the ordering allows the scheduling algorithm to be distributed. We then showed that the ordering as well as the convergence time guarantee generalize to integer edge lengths and capacities.

Seeing how the vertex ordering helped us in obtaining a distributed scheduling algorithm without sacrificing convergence time, we plan to study further implications of this order structure. On the practical side, we hope to put the algorithm onto robots to test its performance in real world applications. With increased availability of cheap and fast wireless communication capabilities, we believe our algorithm can be used on formation control problems for a large group of robots or other types of vehicles in practice.